# An Introductory Review of Spiking Neural Network and Artificial Neural Network: From Biological Intelligence to Artificial Intelligence

Shengjie Zheng
Shenzhen College of Advanced Technology
University of Chinese Academy of Sciences and Shenzhen Institute of Advanced Technology, Chinese Academy of Sciences
Shenzhen, China
sj.zheng@siat.ac.cn

Lang Qian
Tsinghua Shenzhen International Graduate School
Tsinghua University
Shenzhen, China
ql20@mails.tsinghua.edu.cn

Pingsheng Li
Department of Psychology, Department of Biology, and Department Physiology
McGill University
Montreal, Canada
pingsheng.li@mail.mcgill.ca

Chenggang He
Brain Cognition and Brain Disease Institute (BCBDI)
Shenzhen Institute of Advanced Technology, Chinese Academy of Sciences
Shenzhen, China
888hcg@gmail.com

Xiaoqin Qin
School of Information and Communication Engineering
Beijing University of Posts and Telecommunications
Beijing, China
xiaoqiqin@bupt.edu.cn

Xiaojian Li*
BCBDI
Shenzhen Institute of Advanced Technology, Chinese Academy of Sciences
Shenzhen, China
xj.li@siat.ac.cn

*Abstract*—Recently, stemming from the rapid development of artificial intelligence, which has gained expansive success in pattern recognition, robotics, and bioinformatics, neuroscience is also gaining tremendous progress. A kind of spiking neural network with biological interpretability is gradually receiving wide attention, and this kind of neural network is also regarded as one of the directions toward general artificial intelligence. This review introduces the following sections, the biological background of spiking neurons and the theoretical basis, different neuronal models, the connectivity of neural circuits, the mainstream neural network learning mechanisms and network architectures, etc. This review hopes to attract different researchers and advance the development of brain-inspired intelligence and artificial intelligence.

## I. Introduction

Spiking neural networks based on brain-inspired computation, a model that mimics the brain's intelligent computational mechanism, are considered as one of the paths to achieve general artificial intelligence, and this class of algorithms is gaining widespread attention. Meanwhile, traditional deep neural networks have shown extraordinary capabilities in several tasks and seem to have become omnipotent. But there are some key questions: What are the intelligent performances of these two different types of networks? What are the similarities between them?

Inspired by the brain hierarchy and the integration of neural information, artificial neural networks use a multilayer network architecture to transform input information into features, but this precise transformation of data integration is incompatible with the way the brain processes information. Therefore, we discuss here the similarities and differences between biological and artificial intelligence, from the basic information units, network architecture, and learning mechanisms, and how these two types of neural networks represent information as two different ways of processing information.

We first discuss the "biological" aspects, including how neurons integrate information and how information is transmitted between neurons. Then, we explore the biological neuron model and the artificial neuron model, and how these two different types of neurons process information. After that, we discuss how biological neuronal recurrent connected, how different circuits can achieve different functions, and explore the role and embodiment of different neural circuits in biology. Then, we discuss three mainstream machine learning algorithm paradigms and carry out the related biological interpretability discussion, and discuss the advantages and disadvantages of SNN and DNN using different machine learning algorithms. Next, we proceed to discuss the main network architectures of ANNs and SNNs, discuss that SNNs are bio-interpretable, and discuss the importance of network architecture. Finally, we discuss the future of SNNs and ANNs as different types of networks and how they complement each other, including the generality of SNNs and the accuracy of ANNs, to achieve a step towards general artificial intelligence.

## II. BIOLOGICAL BACKGROUND

### A. The Neuron and Synapse

**The Neuron.** The human brain is composed of 86 billion neurons and is the most complex organ within the human body[1]. The highly structured connections between neurons and their interactions form efficient information communication, resulting in neural networks. A classical neuronal structure is composed of three parts: dendrites, soma, and axon. The vast majority of neurons are polarized cells, and cell polarity refers to the spatial variation in shape, structure, and function within a cell. Almost all cell types exhibit some form of polarity, which allows them to perform specialized functions. Dendrites in nerve cells are structured in a dendritic distribution and transmit the received input signals to the soma. With the input of information, the soma changes its own membrane potential in response to all inputs from the dendrites, and when the membrane potential reaches a certain threshold, an action potential is generated. The action potential is transmitted along the axon as an output, and the spike signal is transmitted to the axon terminal.

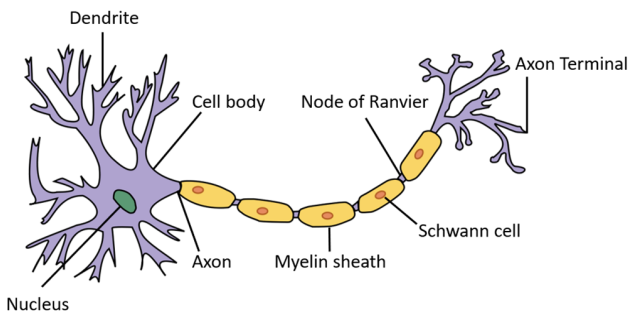

Fig. 1. The Neuron

**The Synapse.** Neurons form connections with each other and transmit information through synapses. The neurons before and after the synapse are the presynaptic and postsynaptic neurons, respectively. For chemical synapses, presynaptic neurons are not directly connected to postsynaptic neurons, but rather to a gap called the synaptic gap. When the action potential travels along the axon to the presynaptic terminal, the presynaptic terminal will produce neurotransmitters, molecules that are packaged into structures called vesicles, and the action potential causes these vesicles to fuse to the membrane and finally be released back into the synaptic gap and bind to receptors on the surface of the postsynaptic terminal, and the binding of different receptors to specific neurotransmitters will affect the changes in the postsynaptic neuron.

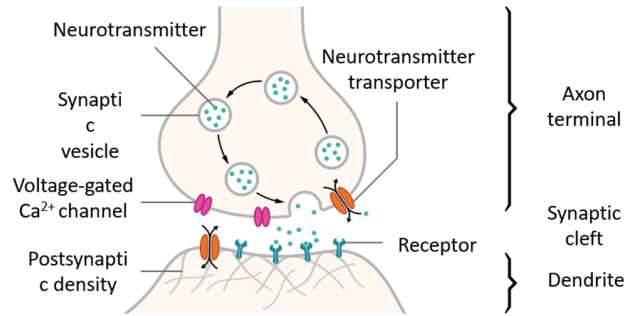

Fig. 2. The Synapse

### B. Action Potential

The membrane potential is the difference in electrical potential between the inner and outer membranes of the neuron, and the key to the generation of action potentials lies with substances inside and outside the cell membrane, generally charged ions and molecules, and we focus here only on charged ions, such as sodium, potassium, calcium, and chloride ions. The cell membrane itself is a good insulator with high electrical resistance but is itself filled with many ion channels that allow ions to flow through the various ion channels. In the cell membrane, there is a very important class of channels called voltage-gated channels, which open or close at any given moment depending on the local membrane potential shift. The membrane potential is initially resting potential, which is the membrane potential of a neuron in the absence of any stimulus. When the neuron is stimulated, the voltage-gated sodium channels will open when the membrane potential reaches a certain stage, causing the membrane potential to rise rapidly, a process called depolarization, until a threshold is reached when the membrane potential rises due to the inactivation of sodium channels and the opening of potassium channels. This process is called depolarization until the threshold is reached, and the membrane potential decreases rapidly due to the inactivation of sodium channels and the opening of potassium channels, which is called hyperpolarization or repolarization, and the membrane potential gradually returns to the resting potential with the closing of potassium channels.

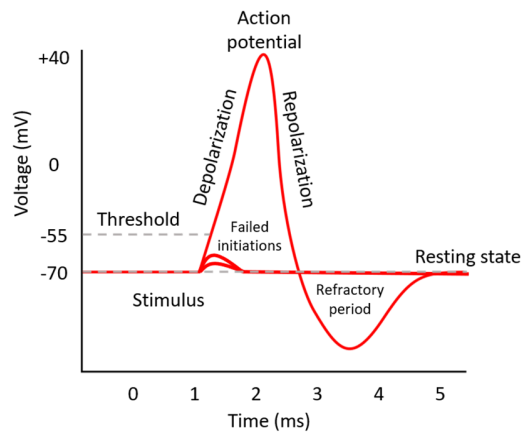

Fig. 3. The Action Potential

## C. Synaptic potential and synaptic integration

Synaptic transmission has two basic forms: excitation and inhibition, and these two form carriers are excitatory neurons and inhibitory neurons, respectively. The signal from an excitatory neuron causes the membrane potential of the postsynaptic neuron to toward a more positive or depolarized value, which prompts the downstream neuron to fire, hence the term excitatory postsynaptic potential (EPSP), for the signal generated by this postsynaptic neuron. Inhibitory neurons cause the membrane potential of the postsynaptic neuron to toward a more negative value. Unlike EPSP, this signal inhibits the downstream neuron in such a way that the membrane potential is farther from the threshold, hence the term inhibitory postsynaptic potential (IPSP), The neuron continuously receives excitatory and inhibitory inputs, and when this input sum reaches or exceeds the threshold, it excites an action potential, otherwise, it remains silent. This process of receiving synaptic inputs is called synaptic integration.

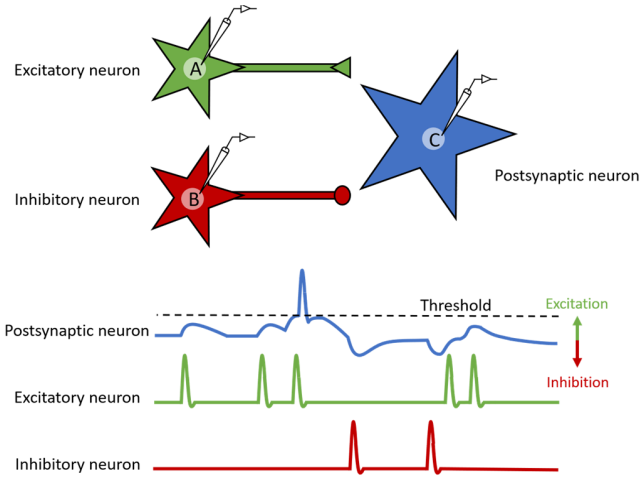

Fig. 4. Synaptic Integration

## III. NEURON MODELS

A neuronal model is a mathematical description of certain properties in the nervous system. Biological neurons are known as action potentials because they generate spike signals that last approximately one millisecond in duration. These spike signals carry extremely strong temporal-spatial properties and contain a large amount of information; therefore, the spiking neurons that generate action potentials are considered information processing units of the nervous system. There are various types of pulse neuron models, such as the Hodgkin and Huxley model, which is based on ion channels and thus simulates membrane potential changes, or the Leaky integrate-and-fire model (LIF) model, which is based directly on stimulus-induced membrane potential changes. A brief description of the basis of the models will be given below.

### A. Hodkin-Huxley model

Hodgkin and Huxley performed experimental recordings on a giant axon of a squid in which they injected electrical currents directly into the axon. A series of careful measurements led to a biophysical model description, an equivalent analog circuit, and a mathematical model. the Hodgkin-Huxley model can be represented by the following differential equation[2].

$$C\frac{du}{dt} = -g_{Na}m^3h(u - E_{Na}) - g_Kn^4(u - E_K) - g_l(u - E_l) + I(t)$$

$$\frac{dm}{dt} = \alpha_m(u)(1 - m) - \beta_m(u)m$$

$$\frac{dn}{dt} = \alpha_n(u)(1 - n) - \beta_n(u)n$$

$$\frac{dh}{dt} = \alpha_h(u)(1 - h) - \beta_h(u)h$$

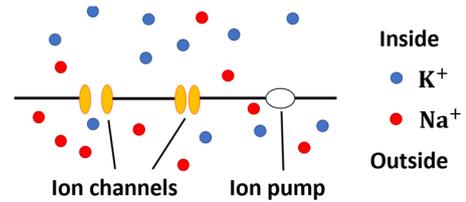

Fig. 5. A brief diagram of the membrane environment

This equivalent simulated circuit corresponds to specific resistances for sodium, potassium, and leak channels, and these resistances change differently over time, with the change in resistance of the simulated circuit corresponding to the opening and closing of the ion channels. Where $I(t)$ represents the input current, $u$ represents the cell membrane potential, $E_{Na}$, $E_K$, and $E_l$ correspond to the reversal potential of sodium, potassium, and leaky channels, $g_{Na}m^3h$, $g_Kn^4$, and $g_l$ correspond to the conductivity of different channels. $m$, $n$, and $h$ are assumed to be the concentrations of certain ion-transport-related particles, and their corresponding $\alpha$ and $\beta$ symbolize the rates of movement of the particles into and out of the membrane. Because of the description of neuronal dynamics at the level of ion channels, this result laid the biophysical foundation of neuroscience, for which Hodgkin and Huxley were awarded the Nobel Prize in 1963.

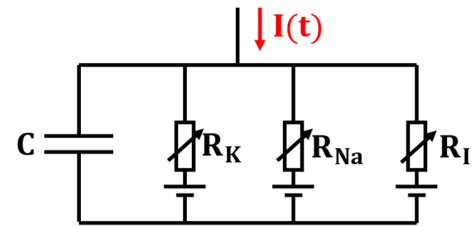

Fig. 6. Hodgkin and Huxley Model

### B. Leaky Integrate and Fire model

The Leaky integrate-and-fire model (LIF), a simplified version of the neuron model, is widely used as the basic unit of spiking neural networks due to its relatively small computational complexity. The basic concept of the LIF neuron was proposed by L.E Lapicque in 1907[3].

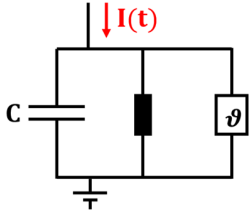

Fig. 7. Leaky Integrate and Fire model

$$\tau \cdot \frac{du}{dt} = -(u - u_{rest}) + R \cdot I(t)$$

In the LIF model, $\tau$ is the time constant of the differential equation and $u_{rest}$ is a constant parameter represented as the resting potential of the cell membrane. $I(t)$ is the input current and R is the membrane resistance.

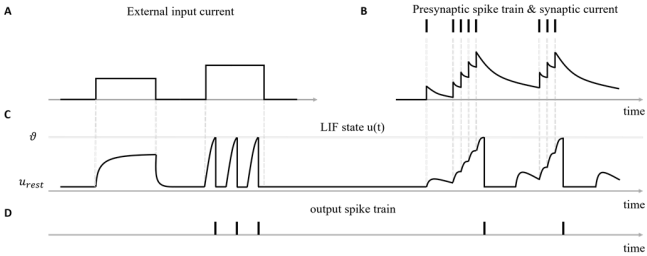

Fig. 8. Time course of the membrane potential of a LIF model driven by the external input signal.

## C. Izhikevich Model

The Izhikevich neuron model is a two-dimensional system of ordinary differential equations[4], as shown below.

$$\frac{dv}{dt} = 0.04v^2 + 5v + 140 - u + I(t)$$

$$\frac{du}{dt} = a(bv - u)$$

where $u$ is a membrane recovery variable used to describe the ion current behavior in general, and $a$, $b$ are used to adjust the timescale of $u$ and the sensitivity about the membrane potential $v$, respectively. By the choice of parameters, the Izhikevich model can demonstrate the firing patterns of almost all known neurons in the cerebral cortex with much less computational overhead than the Hodgkin-Huxley model.

## D. Perceptron

The perceptron[5], as the most basic computational unit in an artificial neural network, perceives a weighted summation of the inputs and then produces output through an activation function, which is essentially a simulation and simplification of a biological neuron. Biological and artificial neurons have many similarities, but biological neurons need to consider the temporal dimension of information as well as the morphological spatial dimension of the neuron itself, while the input and output have a highly nonlinear relationship.

$$f(\mathbf{x}) = \begin{cases} 1 & \text{if } \mathbf{w} \cdot \mathbf{x} + b > 0 \\ 0 & \text{otherwise} \end{cases}$$

$$\mathbf{w} \cdot \mathbf{x} = \sum_{i=1}^{n} w_i x_i = w_1 x_1 + w_2 x_2 + w_3 x_3 + \cdots + w_n x_n$$

The perceptron is defined as a binary classifier, which is actually a mapping function for the input $\mathbf{x}$. $\mathbf{w}$ is the weight value, $\mathbf{w} \cdot \mathbf{x}$ is the dot product, $b$ is the bias, and $\mathbf{w} \cdot \mathbf{x} + b$ is based on the mapping of the binary step function $f(\cdot)$, which yields an output value denoted as "1" or "0", the output value can also be other numbers other number, depending on the chosen activation function $f(\cdot)$.

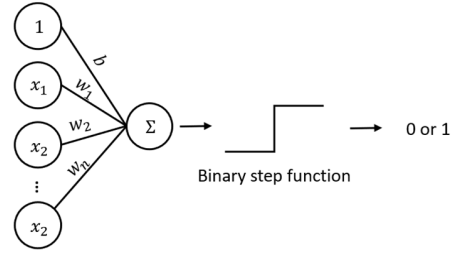

Fig. 9. The perceptron

## E. Relation between artificial and spiking neuron Model

Both neuron models transform input information into output. However, in fact, this is still a different mechanism from that of biological neurons to process information. Biological neurons need to consider the time-dependence about the input and output information which is also the spike information. But it is also the artificial neuron perceptron that chooses a different way of processing information than the biological neuron, together with a specific activation function, which allows the artificial neuron to compute as well as tune the network parameters efficiently in the network model.

## IV. NEURONAL CIRCUIT

Neural circuits are neuron populations interconnected by synapses that perform specific functions when the circuits are activated. The specific way in which these synapses are connected provides the physical basis for neural population dynamics, and these circuits are also used in the architectural design of spiking neural networks. A brief description of neural loops will be given below.

## A. Circuit motifs

Neurons are individual cell units in the nervous system that not only receive input signals from dendrites and process them within the cell body but also send output signals to presynaptic terminals via axons, which neuroanatomist Ramón y Cajal calls "the neuron doctrine"[6].

Neurons do not exist independently within the brain but are highly interconnected in synapses to form circuits that work together to process information, and this connectivity pattern provides the basis for the neuron population to perform specific functions. For example, specific circuits are associated with short-term memory and long-term memory storage, the extent of feature sensory fields, etc. This neuron population, processing information in a similar way to individual neurons, integrate incoming information and then decides whether to perform the output of the information. Also, circuits are modulated by the type of synaptic input they receive, such as excitability as well as inhibition.

*B. Feedforward excitation with Convergence and Divergence*

Feedforward excitation allows a neuron to propagate excitatory signals from itself downstream, and a series of feedforward excitatory connections is common in the nervous system, which allows signals to propagate throughout the system internally, by way of Convergence, where a single postsynaptic neuron receives excitatory signals from multiple presynaptic neurons, and by way of divergence, where a single presynaptic neuron signaling with multiple postsynaptic neurons. Convergent excitation can enable postsynaptic neurons to respond selectively to features not solely or explicitly present in any of the presynaptic neurons. It can also increase the signal-to-noise ratio if multiple input neurons carry the same signal but uncorrelated noise[7].

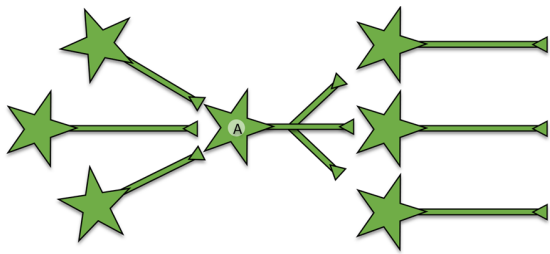

Fig. 10. Feedforward excitation

*C. Feedback/Recurrent excitation*

For Feedback excitation, presynaptic neuron A makes an excitatory connection to postsynaptic neuron B, which in turn connects back to presynaptic neuron A. Moreover, there are also axons of neurons that connect themselves as a recurrent connection.

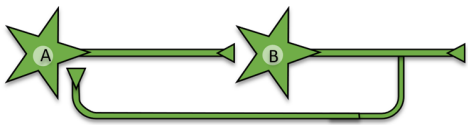

Fig. 11. Feedback/Recurrent excitation

*D. Mutual excitation*

Neuronal cells B, C, D, and E, make excitatory-type interconnections, while neuron B also makes cyclic connections to itself; this excitatory-type connection will allow feedback of excitatory information from the neurons in the network so that the neurons in the network can make prolonged excitatory states corresponding to brief stimuli. This type of connectivity has been used in computational models of working memory[8], as well as short-term memory encoding, which plays an important role.

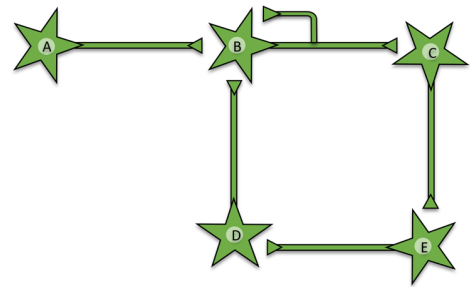

Fig. 12. Mutual excitation

*E. Feedforward inhibition*

When inhibitory neurons are between excitatory neurons, feedforward inhibition is a form of signaling in neuronal transmission, and when presynaptic cells excite the interneuron as inhibitory, the signal from this inhibitory neuron will inhibit the activity of downstream cells.

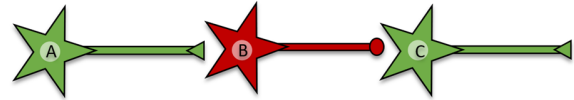

Fig. 13. Feedforward inhibition

*F. Feedback/Recurrent inhibition*

For feedback/Recurrent inhibition, the presynaptic cell is connected to the postsynaptic cell, the postsynaptic cell is connected to the interneuron as an inhibitory effect, and the interneuron is then connected to the presynaptic cell. This circular connection loop serves as a feedback inhibitory effect, which inhibits the activity of excitatory neurons.

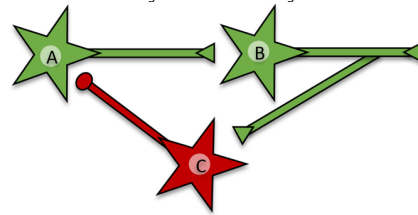

Fig. 14. Feedback/Recurrent inhibition

*G. Lateral inhibition*

Presynaptic cells excite inhibitory interneurons, and thus they inhibit adjacent cells of excitatory neurons. For example, during information processing in the visual system, limbic enhancement is achieved by lateral inhibition of the retina[9, 10].

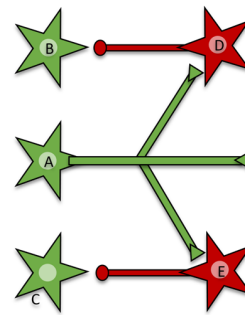

Fig. 15. Lateral inhibition

*H. Mutual inhibition*

Two inhibitory neurons interconnect, when neuron A directly inhibits neuron B, and neuron B receives the inhibitory signal and in turn inhibits neuron A. These mutually inhibitory circuits play a key role in designing the Central Pattern Generator (CPG) computational neural model[11] and in regulating circadian rhythms in the brain[12].

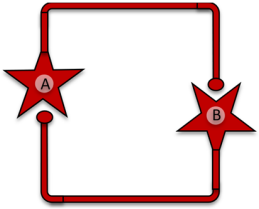

Fig. 16. Mutual inhibition

The interconnection of many neurons in a neural system results in changes in both structure and function as the size increases, and such changes have the emergent properties of physical systems, which arise from the interaction of different elements in the system. Currently, artificial neural networks and spiking neural networks are also trying to make neural networks with emergent properties, or intelligent emergent properties, through different network architectures and neural circuit designs.

## V. LEARNING TYPES

In machine learning for artificial intelligence, there are three main types of machine learning, namely supervised learning, unsupervised learning, and reinforcement learning, which are also used in artificial neural networks as well as spiking neural networks. These learning methods allow neural networks to learn based on data and thus tune the internal parameters of the network to achieve the ability to solve tasks.

*A. Supervised learning*

In general, supervised learning deals with labeled data, which is a process of finding a mapping from the input to the output space. The tasks of learning can be divided into two categories, classification, and regression. The tasks need to produce the desired output for each input. This learning mechanism has the advantage of being able to obtain patterns in the data based on prior knowledge, but only if high-quality data inputs, as well as corresponding outputs, are required. Therefore, it is one of the types of learning that can achieve efficient learning tasks and obtain the best solutions.

**Supervised learning in ANN.**

Artificial neural networks are interconnected through multiple layers of nodes, and such neural networks can be learned in a supervised method so as to find a mapping through a loss function, sometimes called an error function, which indicates a function of the distance between the current output and the expected output[13]. The network parameters are adjusted according to the loss function by means of gradient descent. The model gives the best answer when the loss function is close to zero.

Gradient descent and backpropagation often occur together, and backpropagation is an algorithm for training ANNs for supervised learning. It efficiently calculates the gradient of the error function with respect to the network weights of a single input-output example. Thus, the weights are updated to minimize the loss function. In 1986, Seppo Linnainmaa proposed automatic differentiation. It was later used in experiments for learning internal representations by D.E Rumelhart et al[14]. The successful application of backpropagation has caused a renaissance in the field of research on artificial neural networks. Today, thanks to powerful GPU computing systems, BP algorithms show great advantages in the training of different networks.

**Supervised learning in ANN.**

In traditional artificial neural networks, labels can be represented as integers (classification) or real numbers (regression). In spiking neural networks, the labels are encoded as spike trains with spatial-temporal properties. However, unlike artificial neural networks, it is difficult and unwise to directly apply gradient descent methods to SNNs due to the discontinuity of spike signals.

(1) **Spike Response Model (SRM).** The first gradient descent-based supervised learning algorithm for SNNs was proposed by Bohte et al[15]. The method implements a gradient descent-based multilayer spiking neural network error backpropagation method using the temporal encoding of spike intervals.
(2) **Spike Pattern Association Neuron (SPAN).** Another very powerful supervised algorithm is SPAN[16], whose core idea is to obtain an analog signal by convolving our chosen kernel function with the spike signal, where the Widrow-Hoff rule (Delta rule) can be directly applied to the transformed signal to adjust the weights in the network
(3) **Surrogate Gradient Descent.** Recently, the Surrogate gradient descent was proposed by Zenke et al[17]. It introduces continuous relaxation of gradient estimation without affecting the forward transmission of spike sequences to achieve very good results.

Supervised learning requires annotation of the raw data, a process that is labor-intensive, and data annotation is the foundation of most artificial intelligence, which determines the quality of deep learning models. However, this differs from biological intelligence in that human beings often do not have a precise annotation during the learning process, and that the parameters of the entire network are updated based on backpropagation in a way that does not exist within biological structures. Thus, supervised learning is different from biological learning.

*B. Unsupervised learning*

Unsupervised learning is an algorithm for learning patterns from unlabeled data, which is important in the biological learning process. At the same time, this learning approach hopes to solve the problem of supervised learning requiring data labeling, because, in real life, it is difficult for labelers to perform manual labeling if they lack sufficient a priori knowledge. The development of unsupervised learning is solving these problems and has even rivaled supervised learning in uncovering hidden structures in the data.

**Unsupervised learning in ANN.**

Unsupervised learning based on artificial neural networks has gradually become known as a research hotspot in recent years, and has even been called the next stop for AI. Yann LeCun once used a cake analogy, "If intelligence is a piece of cake, then most of the cake is unsupervised learning, the frosting on the cake is supervised learning, and the cherry on the cake is reinforcement learning". Indeed, unsupervised learning with great potential is already producing significant results in several fields.

(1) **Autoencoder[18].** It is an unsupervised neural network model in artificial neural networks that learns the implicit features of the input data, which is called encoding while reconstructing the original input data with the learned new features, called decoding. Autoencoder can be used for feature extraction as well as dimensionality reduction.
(2) **Generative adversarial network (GAN)[19].** Generative adversarial networks learn by letting two networks contest each other, consisting of a generative network as well as a discriminative network, respectively. The generative network takes random samples from the latent space as input, and the output needs to be similar to the real samples of the training set. The purpose of the discriminative network is to be able to discriminate the output of the generative network.
(3) **Self-organizing map (SOM)[20].** A self-organizing mapping neural network, which uses unsupervised learning to produce a low-dimensional representation of the input space of discretized training samples, is called mapping and is, therefore, a method of dimensionality reduction. Self-organizing mapping differs from other artificial neural networks because it uses competitive learning in which output neurons compete with each other for activation, with only one neuron being activated at any given time, called winner-takes-all neuron.

**Unsupervised learning in SNN.**

In the process of biological learning, we know that even when we are in infancy, we are able to achieve recognition as well as comparison of things that are not labeled. This seems to indicate that biological neural networks are capable of unsupervised learning of external input information to build an initial understanding of the outside world. The following will describe the use of spiking neurons to model neural networks using biologically based explanatory learning rules to build the basis for biological learning and memory.

(1) **Hebbian learning.** During learning and memory, the weight of synaptic connections between biological neurons is strengthened or weakened in response to the activity between neurons, which is crucial for information storage in the brain. Donald Hebb assumes that the persistence or repetition of a reverberatory activity tends to induce lasting cellular changes that add to its stability[21]. This theory is also known as Hebb's rule, Hebb's postulate, and cell assembly theory. Hebb's rule is also summarized as "fire together wire together".
(2) **Spike-Timing Dependent Plasticity (STDP).** STDP refers to the observation that the precise spike timing influences the enhancement and inhibition of synaptic plasticity[22]. For example, in the connections between mammalian pyramidal neurons, when presynaptic spikes occur within a certain time window of postsynaptic spikes, it will lead to long-time-travel potentiation (LTP); if the order is reversed, it will lead to long-time-travel depression (LTD), a phenomenon that occurs in the synapses of biological neurons, although it is not yet applicable to all brain regions and cell types.

$$\Delta w = \begin{cases} a^+ e^{\frac{-(|t_{pre}-t_{post}|)}{\tau}} & t_{pre} - t_{post} \leq 0, a^+ > 0 \\ a^- e^{\frac{-(|t_{pre}-t_{post}|)}{\tau}} & t_{pre} - t_{post} > 0, a^- < 0 \end{cases}$$

$\Delta w$ represents the amount of synaptic weight change, $a^+$ and $a^-$ are the learning rate parameters, respectively, $\tau$ is the time constant, and $t_{pre} - t_{post}$ represents the time difference between presynaptic neuron spike delivery and postsynaptic neuron spike delivery, respectively

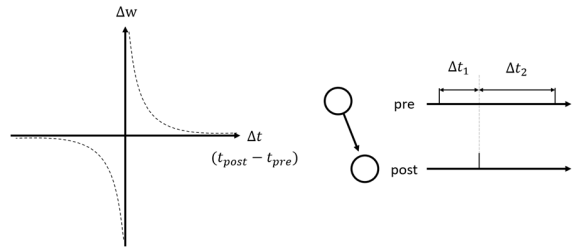

Fig. 17. Spike-Timing Dependent Plasticity

(3) **Triplets STPD.** In 2006, Triplet was proposed by Pfister and Gerstner[23], in which LTP is constructed as a combination of one presynaptic as well as two postsynaptic spikes, and LTD is based on the combination of two presynaptic as well as one postsynaptic spike instead of a pair of spikes based on STDP rule, this rule can well take into account the spike-timing interactions, but still also does not apply to the interpretation of all cell types.

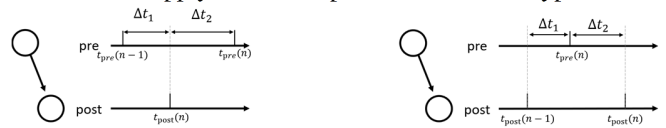

Fig. 18. The triplets STDP

For now, biologically interpretable learning rules for unsupervised learning are still to be explored, especially for SNNs, based on the fact that unsupervised learning still has a large gap for complex tasks compared with supervised learning, and the main reason in this regard is still based on the fact that current unsupervised learning rules are not perfect enough to enable spiking neural networks to learn from data effectively. More explanatory unsupervised learning algorithms will be the key to achieve SNN performance improvement.

*C. Reinforcement learning*

Reinforcement learning (RL) is a machine learning approach to artificial intelligence that works to create computer programs capable of solving problems that require intelligence. This learning algorithm is inspired by the study of reward mechanisms for animal learning.

**Reinforcement learning in ANN.**

The unique feature of RL based on artificial neural networks is that it learns from feedback through iterative trials that are simultaneously sequential and evaluable through the use of powerful nonlinear function approximations. As a framework for solving control tasks, it learns from the environment by constructing agents that interact with the environment through iterative attempts and receive rewards as the only feedback.

(1) **Value-based.** Learn the state or state-action value. Act by choosing the best action to take in this state. Q-learning is one of the most classic value-based algorithms[24], the Deep Q-Network algorithm, which was proposed by DeepMind in 2015[25], enables algorithms to play Atari games like humans by combining Q-Learning in reinforcement learning and deep neural networks. They accept several frames of the game as an input and take the output state value of each action as output.
(2) **Policy-based.** As REINFORCE Gradient from Williams[26], the policy is learned as a mapping from the state space to the action space, telling the agent the best action to take in each state to maximize its return.

**Reinforcement learning in SNN.**

In the process of biological learning, many learning processes are accompanied by reward mechanisms. This appears to be a global reinforcement signal acting on multiple regions of the brain, resulting in changes in the connectivity of the neural network. And spiking neural networks have great potential to mimic this way of biological learning, adjusting their own network connections to get the most out of the reward signal.

(1) **Three-factor Learning Rules.** This approach works by setting a flag, called an eligibility trace[27], on the synapse upon co-activation of presynaptic and postsynaptic neurons. synaptic weights change only when a third factor, indicating reward, is present when the flag is set.
(2) **ANN to SNN.** By matching the firing frequency of the firing neurons and the graded activation of the analog neurons, the trained artificial neural networks can be converted to the corresponding SNNs[28]

Reinforcement learning is extremely biologically interpretable, a theory inspired by the psychology of how agents gradually develop expectations of stimuli in response to rewarding or punishing stimuli given by the environment, producing habitual behaviors that yield the greatest benefit. There is a prevailing view that dopamine neurons enable this function, comparing future expectations with previous mental benchmarks and thus releasing neurotransmitters depending on the result, thus making the creature happy or frustrated, using a reward mechanism as the basis for learning[29]. However, current reinforcement learning using deep learning in practice generally requires a large amount of training time, and how to truly learn based on reward and efficiently like a living creature is something that currently needs to be improved.

## VI. Architectures of Neural Network

ANN artificial neural networks are inspired by the brain, but compared to the brain, there are fundamental differences in both neuronal topology, neural information processing, and neural learning mechanisms. Within the biological context, neurons communicate with each other by passing spikes, and the information is embedded in the spike trains. ANNs use continuous value as the information transmitted between artificial neurons, and the model architecture is of the human-designed type, which generally has no biological functional properties. With the rapid development of deep artificial neural networks and neuromorphic hardware, these advances have simultaneously led to new research and hypotheses on spiking neural networks.

*A. Feedforward Neural Network*

**Artificial Feedforward Neural Network.**

Feedforward Neural Network, also known as Multilayer perceptron (MLP)[30], is essentially a nonlinear composite function $N(\cdot)$ that approximates a certain function $y(\cdot)$, which maps the input x to the output $y(x)$. The input x is able to obtain the corresponding output through a series of nonlinear transformations $f^{(k)}(\cdot)$, i.e., the nonlinear activation function $\sigma(\cdot)$. The input is passed forward after the nonlinear transformation in each layer, and finally the output is obtained. Usually, the first layer of the network is called the input layer, the last layer is called the output layer, and the one between the first and the last layer is called the hidden layer, and each unit in the network is also known as the perceptron. As the number of hidden layers of the network increases and the number of units within the hidden layers increases, the complexity of the FNN nonlinear model $N(x)$ increases, so it can approximate any function that can satisfy any nonlinear mapping relationship between the input x and the output $y(x)$.

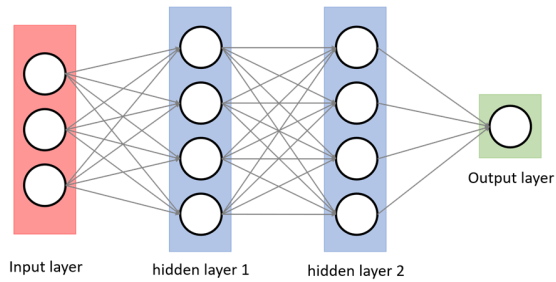

Fig. 19. Feedforward Neural Network

**Artificial Feedforward Neural Network.**
There are many SNNs based on STDP learning or BP-based supervised learning that have achieved success in different types of pattern recognition, and even some of the STDP-based networks are comparable to BP-based supervised learning, such as Diehl et al[31], who showed that using a two-layer SNN, based on the biological properties of excitatory type neurons and inhibitory neurons as the processing layer, using lateral inhibition as well as winner-take-all properties, enabling the neurons in the processing layer to extract features with significant characteristics from the input signal based on STDP learning rules, with optimal performance of 95% on the MNIST dataset.

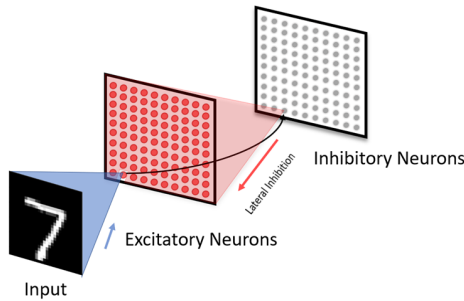

Fig. 20. Spiking Feedforward Neural Network with Lateral Inhibition

### B. Convolutional Neural Network

**Artificial Convolutional Neural Network.**
CNN is usually used to process gridded data (e.g., images) and consists of layers that process visual information, the most commonly used layers being convolutional, pooling, and fully connected layers[32]. CNN learns the spatial patterns in an image region by looking at groups of pixels in it. the convolutional layer looks for spatial features from the input to perform feature extraction, and this operation is performed through a series of filters, also known as convolutional kernels, convolves the input, which is essentially a cross-correlation operation, followed by a nonlinear activation function, and multiple filters to obtain multiple corresponding outputs. The convolution layer is followed by the pooling layer, which reduces the size of the feature space to reduce the number of parameters in the network and the amount of computation, while it helps to extract dominant features of translation invariance, thus making the model training effective. The final layer is the fully-connected layer, where the initial input is transformed into highly abstract and low-dimensional information representing the input through a series of convolutional and pooling layers. The fully connected layer transforms the output of the feature extraction layer into a vector and connects the feedforward neural network FNN, and the last layer classifies the output by a softmax function.

The convolutional layers as well as the pooling layers in CNN come from the concept of simple and complex cells in neuroscience[33]. The overall architecture of CNN has some similarities with the visual ventral pathway LGN-V1-V2-V4-IT, for example, the layers near the input may represent the contour information of a picture, while the layers near the output can be more representative of the category.

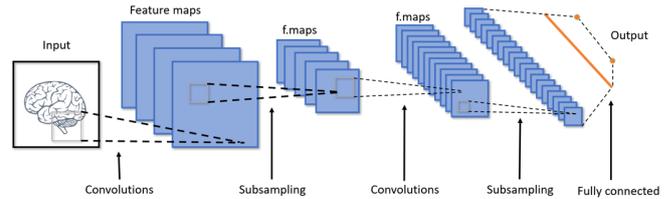

Fig. 21. Convolutional Neural Network

**Spiking Convolutional Neural Network.**
There are convolutional kernels that use V1-like properties for the CNN closest to the input level, which can extract salient features for images. For example, SR Kheradpisheh et al. used a Difference-of-Gaussian kernel[34] for the input image, followed by unsupervised STDP-based training of the convolutional layer as well as the pooling layer, and finally, the extracted features are passed into the classifier[35]. The performance of directly trained SNNs is often inferior to that of traditional DNNs, while training on non-neuromorphic hardware is time-consuming, and ANN conversion to SNNs can solve this problem. Many studies have shown that converted SCNNs work well and perform close to CNN and that SCNNs can perform inference tasks on neuromorphic hardware[36] and consume less energy.

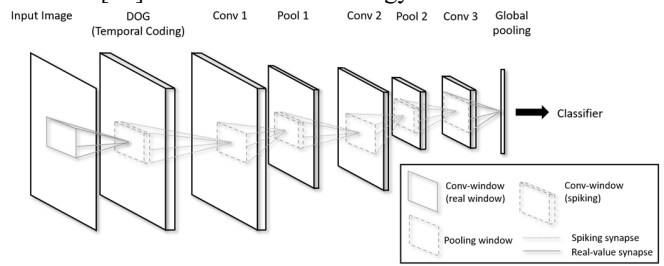

Fig. 22. Spiking Convolutional Neural Network

### C. Recurrent Neural Network

**Artificial Recurrent Neural Network.**
RNN is a class of neural networks used to process sequential data or time-series data, often used to solve ordinal or temporal problems, such as language translation, speech recognition[37], etc. RNNs, like FNNs and CNNs, are also composed of the difference that lies in the learning process. Instead of memorizing the overall sequence information, the RNN uses the representational information in the hidden layer to memorize the information in the most recent time step in the learning process based on time series, and the RNN combines the representational information of the previous time step in the hidden layer with the input of the current step to infer the output of the current time step. Currently, RNN-based Gated recurrent unit (GRU)[38] as well as Long short-term memory (LSTM)[39] have been used to powerful effect in real-life applications.

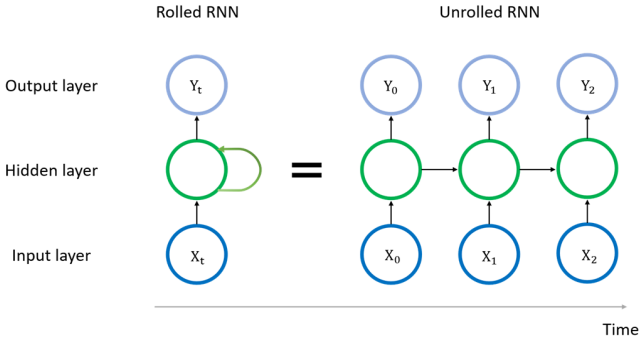

Fig. 23. Recurrent Neural Network

**Spiking Recurrent Neural Network.**

Neural circuits in the brain display the remarkable dynamic richness and high variability in the form of recurrent connections. Excitatory and inhibitory neurons interconnect to form a neural network that is in a chaotic as well as an equilibrium state transition. This recurrent neural network has complex nonlinear dynamics and can be used to study biological neural networks in specific microcircuits of the brain. People often use Liquid state machines (LSM) for computational modeling, the essence of LSM is related to its own naming, the idea is to throw a stone into a lake, the lake indicates that ripples will be generated, based on the current activity of the lake, it is possible to evaluate what happened previously in the system, such as how long ago the stone entered the lake and thus caused the ripples[40]. Essentially, the lake is the LSM, the rocks are the input, and the ripples are the cluster response of the LSM. the LSM usually consists of three layers, the input layer, the reservoir or liquid layer, and the memoryless readout layer. this recurrent neural network transforms the time-varying input information into a higher dimensional space by it can exhibit rich temporal as well as spatial properties of neuronal dynamics, and thus can memorize past input information.

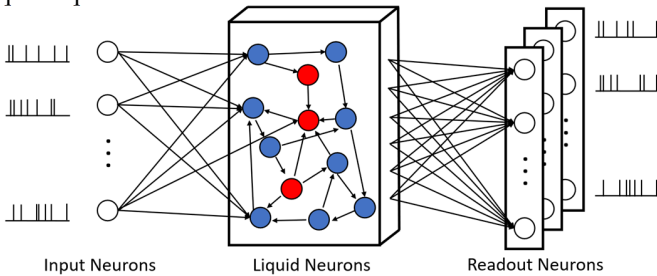

Fig. 24. Liquid state machines

The emergence of these different network architectures actually comes from the understanding of the neural system. How to implement different tasks in a generic architecture cannot yet be done on ANNs as well as SNNs. Nevertheless, in many ways, SNNs and ANNs can be complementary and do not replace each other, and for brain science and brain-inspired research, SNNs are of great importance. Faced with computationally oriented tasks, ANNs have unparalleled advantages. Currently, a class of network models combining SNN and ANN has been born that can exploit the advantages of each. With the understanding of biological neural networks, it should be possible to inspire new neural network architectures, and through the conditioning of large-scale neural information as well as recording, it is possible to further understand the patterns of neural information within the network, providing a basis for neural network architectures with biological explanations.

VII. CONCLUSION

In this paper, we first analyze the biological background of neurons and then describe mathematical models of biological neurons that simulate the changes in membrane potential of neurons with different computational complexity. An overview of neural circuits in biology is also given, with different circuits implementing different information processing functions. This is followed by an application of the mainstream learning mechanisms in different neural networks. Finally, a review of the mainstream network architectures is presented.

The review shows that the way neurons are connected and the learning mechanisms are the basis of brain learning and that for different network architectures and different learning mechanisms, neural networks will exhibit different information processing. For SNNs, the use of a bio-interpretive STDP learning mechanism and a recurrent connection network structure have excellent advantages for processing spatial-temporal information. For ANNs, this artificially designed structure with set learning mechanisms and high-quality big data can lead to models that explain the hidden structural patterns of the data.

However, current neural networks face some potential challenges, at least for both SNNs and ANNs. for example, it is unclear whether neural circuits simulated using SNNs can explain biological neural circuits, and the use of different parameters is likely to yield different results. Also, although ANNs can work well in principle and have been used very successfully in engineering, in order to achieve specific functions, ANNs need to receive specific data as well as learning method constraints, which are often different from biological processes. Although more and more ANN models have recently applied new learning paradigms, ANNs currently fail to achieve better results for multimodal tasks, as they are still inherently data-driven. How to learn effectively with small-scale data will be the key to the progress of ANN or SNN, because they can effectively reveal the nature of learning directly based on the inherent prior knowledge of the network and can evaluate the effectiveness of data learning.

Perhaps, in the future, neural microcircuits in the brain and large-scale neuronal cluster acquisition and analysis can provide new tools that will play a crucial role in discovering neural network architectures as well as learning mechanisms. There may also be a need to integrate advances in different fields, such as neuromorphic chips and hybrid chips, which will facilitate the development of AI as well as brain-inspired intelligence at different levels.

Finally, the neural networks based on general AI and brain-inspired intelligence are far from complete. For example, what computations are done by the dendrites of individual neurons on the inputs[41], what is the internal information flow process of the neural circuits of biological neural networks, and the uninterpretability of backpropagation algorithms in biological neural networks. Whether ANN can be combined with current biological laws to achieve similar effects of biological neural networks.

If the neural network gets the further breakthrough, it may be possible to expose the nature of neural networks for information encoding from another perspective and can effectively solve some problems such as learning and memory, motor planning, pattern recognition, etc. In a new theoretical architecture, either ANN or SNN, we may be able to look at data with a new perspective and perhaps explain the theory of artificial intelligence or brain-inspired intelligence in another way.

ACKNOWLEDGMENT

This work was supported by Key Area R&D Program of Guangdong Province with grant No. 2018B030338001.

REFERENCES

[1] S. Herculano-Houzel, "The remarkable, yet not extraordinary, human brain as a scaled-up primate brain and its associated cost," *Proceedings of the National Academy of Sciences,* vol. 109, no. Supplement 1, pp. 10661-10668, 2012, doi: 10.1073/pnas.1201895109.

[2] A. L. Hodgkin and A. F. Huxley, "A quantitative description of membrane current and its application to conduction and excitation in nerve," *J Physiol,* vol. 117, no. 4, pp. 500-44, Aug 1952, doi: 10.1113/jphysiol.1952.sp004764.

[3] L. Lapique, "Recherches quantitatives sur l'excitation electrique des nerfs traitee comme une polarization," *Journal of Physiology and Pathololgy,* vol. 9, pp. 620-635, 1907.

[4] E. M. Izhikevich, "Simple model of spiking neurons," *IEEE Trans Neural Netw,* vol. 14, no. 6, pp. 1569-72, 2003, doi: 10.1109/TNN.2003.820440.

[5] F. Rosenblatt, "The Perceptron - a Probabilistic Model for Information-Storage and Organization in the Brain," (in English), *Psychological Review,* vol. 65, no. 6, pp. 386-408, 1958, doi: DOI 10.1037/h0042519.

[6] S. R. y Cajal, *Estructura de los centros nerviosos de las aves*. 1888.

[7] L. Luo, "Architectures of neuronal circuits," *Science,* vol. 373, no. 6559, p. eabg7285, 2021, doi: doi:10.1126/science.abg7285.

[8] D. Durstewitz, J. K. Seamans, and T. J. Sejnowski, "Neurocomputational models of working memory," *Nat Neurosci,* vol. 3 Suppl, pp. 1184-91, Nov 2000, doi: 10.1038/81460.

[9] S. W. Kuffler, "Discharge patterns and functional organization of mammalian retina," *Journal of neurophysiology,* vol. 16, no. 1, pp. 37-68, 1953.

[10] H. B. Barlow, "Summation and inhibition in the frog's retina," *The Journal of physiology,* vol. 119, no. 1, pp. 69-88, 1953.

[11] P. A. Guertin, "The mammalian central pattern generator for locomotion," *Brain research reviews,* vol. 62, no. 1, pp. 45-56, 2009.

[12] M. H. Hastings, A. B. Reddy, and E. S. Maywood, "A clockwork web: circadian timing in brain and periphery, in health and disease," (in eng), *Nat Rev Neurosci,* vol. 4, no. 8, pp. 649-61, Aug 2003, doi: 10.1038/nrn1177.

[13] V. K. Ojha, A. Abraham, and V. Snasel, "Metaheuristic design of feedforward neural networks: A review of two decades of research," (in English), *Eng Appl Artif Intel,* vol. 60, pp. 97-116, Apr 2017, doi: 10.1016/j.engappai.2017.01.013.

[14] D. E. Rumelhart, G. E. Hinton, and R. J. Williams, "Learning Representations by Back-Propagating Errors," (in English), *Nature,* vol. 323, no. 6088, pp. 533-536, Oct 9 1986, doi: DOI 10.1038/323533a0.

[15] S. M. Bohte, J. N. Kok, and H. La Poutre, "Error-backpropagation in temporally encoded networks of spiking neurons," (in English), *Neurocomputing,* vol. 48, pp. 17-37, Oct 2002, doi: Pii S0925-2312(01)00658-0
Doi 10.1016/S0925-2312(01)00658-0.

[16] A. Mohemmed, S. Schliebs, S. Matsuda, and N. Kasabov, "Span: Spike Pattern Association Neuron for Learning Spatio-Temporal Spike Patterns," (in English), *Int J Neural Syst,* vol. 22, no. 4, Aug 2012, doi: Artn 1250012
10.1142/S0129065712500128.

[17] E. O. Neftci, H. Mostafa, and F. Zenke, "Surrogate Gradient Learning in Spiking Neural Networks: Bringing the Power of Gradient-based optimization to spiking neural networks," (in English), *Ieee Signal Proc Mag,* vol. 36, no. 6, pp. 51-63, Nov 2019, doi: 10.1109/Msp.2019.2931595.

[18] M. A. Kramer, "Nonlinear Principal Component Analysis Using Autoassociative Neural Networks," (in English), *Aiche J,* vol. 37, no. 2, pp. 233-243, Feb 1991, doi: DOI 10.1002/aic.690370209.

[19] I. J. Goodfellow *et al.*, "Generative Adversarial Nets," (in English), *Advances in Neural Information Processing Systems 27 (Nips 2014),* vol. 27, pp. 2672-2680, 2014. [Online]. Available: <Go to ISI>://WOS:000452647101094.

[20] T. Kohonen, "Self-Organized Formation of Topologically Correct Feature Maps," (in English), *Biological Cybernetics,* vol. 43, no. 1, pp. 59-69, 1982, doi: Doi 10.1007/Bf00337288.

[21] D. O. Hebb, *The organization of behavior; a neuropsychological theory* (The organization of behavior; a neuropsychological theory.). Oxford, England: Wiley, 1949, pp. xix, 335-xix, 335.

[22] G. Q. Bi and M. M. Poo, "Synaptic modifications in cultured hippocampal neurons: Dependence on spike timing, synaptic strength, and postsynaptic cell type," (in English), *Journal of Neuroscience,* vol. 18, no. 24, pp. 10464-10472, Dec 15 1998, doi: DOI 10.1523/jneurosci.18-24-10464.1998.

[23] J. P. Pfister and W. Gerstner, "Triplets of spikes in a model of spike timing-dependent plasticity," *J Neurosci,* vol. 26, no. 38, pp. 9673-82, Sep 20 2006, doi: 10.1523/JNEUROSCI.1425-06.2006.

[24] C. J. Watkins and P. Dayan, "Q-learning," *Mach Learn,* vol. 8, no. 3, pp. 279-292, 1992.

[25] V. Mnih *et al.*, "Human-level control through deep reinforcement learning," (in English), *Nature,* vol. 518, no. 7540, pp. 529-533, Feb 26 2015, doi: 10.1038/nature14236.

[26] R. J. Williams, "Simple Statistical Gradient-Following Algorithms for Connectionist Reinforcement Learning," (in English), *Mach Learn,* vol. 8, no. 3-4, pp. 229-256, May 1992, doi: Doi 10.1023/A:1022672621406.

[27] N. Fremaux and W. Gerstner, "Neuromodulated Spike-Timing-Dependent Plasticity, and Theory of Three-Factor Learning Rules," *Front Neural Circuits,* vol. 9, p. 85, 2015, doi: 10.3389/fncir.2015.00085.

[28] B. Rueckauer, I. A. Lungu, Y. Hu, M. Pfeiffer, and S. C. Liu, "Conversion of Continuous-Valued Deep Networks to Efficient Event-Driven Networks for Image Classification," *Front Neurosci,* vol. 11, p. 682, 2017, doi: 10.3389/fnins.2017.00682.

[29] Y. Niv, M. O. Duff, and P. Dayan, "Dopamine, uncertainty and TD learning," *Behav Brain Funct,* vol. 1, p. 6, May 4 2005, doi: 10.1186/1744-9081-1-6.

[30] Y. LeCun, Y. Bengio, and G. Hinton, "Deep learning," (in English), *Nature,* vol. 521, no. 7553, pp. 436-444, May 28 2015, doi: 10.1038/nature14539.

[31] P. U. Diehl and M. Cook, "Unsupervised learning of digit recognition using spike-timing-dependent plasticity," (in English), *Frontiers in Computational Neuroscience,* vol. 9, Aug 3 2015, doi: ARTN 99
10.3389/fncom.2015.00099.

[32] Y. Lecun, L. Bottou, Y. Bengio, and P. Haffner, "Gradient-based learning applied to document recognition," (in English), *P Ieee,* vol. 86, no. 11, pp. 2278-2324, Nov 1998, doi: Doi 10.1109/5.726791.

[33] Y. LeCun and Y. Bengio, "Convolutional networks for images, speech, and time series," *The handbook of brain theory and neural networks,* vol. 3361, no. 10, p. 1995, 1995.

[34] D. Marr and E. Hildreth, "Theory of edge detection," *Proceedings of the Royal Society of London. Series B. Biological Sciences,* vol. 207, no. 1167, pp. 187-217, 1980.

[35] S. R. Kheradpisheh, M. Ganjtabesh, S. J. Thorpe, and T. Masquelier, "STDP-based spiking deep convolutional neural networks for object recognition," (in English), *Neural Networks,* vol. 99, pp. 56-67, Mar 2018, doi: 10.1016/j.neunet.2017.12.005.

[36] C. Mead, "Neuromorphic electronic systems," *P Ieee,* vol. 78, no. 10, pp. 1629-1636, 1990.

[37] H. Sak, A. Senior, and F. Beaufays, "Long Short-Term Memory Recurrent Neural Network Architectures for Large Scale Acoustic Modeling," (in English), *Interspeech,* pp. 338-342, 2014. [Online]. Available: <Go to ISI>://WOS:000395050100069.

[38] K. Cho, B. Van Merriënboer, D. Bahdanau, and Y. Bengio, "On the properties of neural machine translation: Encoder-decoder approaches," *arXiv preprint arXiv:1409.1259,* 2014.

[39] S. Hochreiter and J. Schmidhuber, "Long short-term memory," *Neural Comput,* vol. 9, no. 8, pp. 1735-80, Nov 15 1997, doi: 10.1162/neco.1997.9.8.1735.

[40] W. Maass and H. Markram, "On the computational power of circuits of spiking neurons," *Journal of computer and system sciences,* vol. 69, no. 4, pp. 593-616, 2004.

[41] R. Yuste, "From the neuron doctrine to neural networks," (in English), *Nature Reviews Neuroscience,* vol. 16, no. 8, pp. 487-497, Aug 2015, doi: 10.1038/nrn3962

# Authors' Background

| Name | Email | Position (Prof, Assoc. Prof. etc.) | Research Field | Homepage URL |
|---|---|---|---|---|
| Shengjie Zheng | sj.zheng@siat.ac.cn | Graduate student | Intracortical Brain Computer Interface and Brain-inspired intelligence | |
| Lang Qian | ql20@mails.tsinghua.edu.cn | Graduate student | Artificial Intelligence | |
| Pingsheng Li | pingsheng.li@mail.mcgill.ca | Undergraduate student | Computational Neuroscience | |
| Chenggang He | 888hcg@gmail.com | Research assistant | Artificial Intelligence | |
| Xiaoqin Qin | xiaoqiqin@bupt.edu.cn | Associate Professor | Next generation mobile communication basic theory and performance analysis | https://teacher.bupt.edu.cn/qinxiaoqi/zh_CN/index/75333/list/index.htm |
| Xiaojian Li | xj.li@siat.ac.cn | Professor of Engineering | Brain neural information decoding and functional neural circuit analysis; high-performance implantable brain-computer interface systems; nanotechnology-based neuromodulation methods, flexible micro and nano neural interface device design; neural simulation computing and brain-inspired device design, etc. | http://people.ucas.ac.cn/~Lixiaojian https://www.x-mol.com/groups/li_xiaojian |